\documentclass{article}

\usepackage[final]{corl_2018} % Uncomment for the camera-ready ``final'' version

\usepackage{times}
\usepackage{amsmath}
\usepackage{amssymb}
\usepackage{capt-of}
\usepackage{graphicx}
\usepackage{latexsym}
\usepackage{float}
\usepackage{tabularx}
\usepackage{subcaption}
\usepackage[toc,page]{appendix}
\usepackage[algoruled,linesnumbered,noend]{algorithm2e}
\usepackage[skip=5pt]{caption}
\graphicspath{ {./img/} }

\title{Interpretable Latent Spaces for\\ Learning from Demonstration}

\author{
  Yordan Hristov\\
  University of Edinburgh \\
  \texttt{yordan.hristov@ed.ac.uk} \\
  \And
  Alex Lascarides \\
  University of Edinburgh \\
  \texttt{alex@inf.ed.ac.uk} \\
  \And
  Subramanian Ramamoorthy \\
  University of Edinburgh \\
  \texttt{sramamoo@inf.ed.ac.uk} \\
}

\begin{document}
\maketitle

%===============================================================================

\begin{abstract}
Effective human-robot interaction, such as in robot learning from human demonstration, requires the learning agent to be able to ground abstract concepts (such as those contained within instructions) in a corresponding high-dimensional sensory input stream from the world. Models such as deep neural networks, with high capacity through their large parameter spaces, can be used to compress the high-dimensional sensory data to lower dimensional representations. These low-dimensional representations facilitate symbol grounding, but may not guarantee that the representation would be human-interpretable. We propose a method which utilises the grouping of user-defined symbols and their corresponding sensory observations in order to align the learnt compressed latent representation with the semantic notions contained in the abstract labels. We demonstrate this through experiments with both simulated and real-world object data, showing that such alignment can be achieved in a process of physical symbol grounding.
\end{abstract}


%===============================================================================

\section{Introduction}

We want our autonomous robots to be competent across broad domains. Ideally, they must be able to cope with the open world, i.e., one where it may not be possible to provide a complete specification of the task up front. Thus, it is important for the robotic agent to incrementally learn about the world and its regularities in order to be able to efficiently generalize to new situations. It can be argued that such generalizations are made easier when the models used by the robot can represent concepts ranging from object-hood to causality. 

Learning from human demonstration \citep{argall2009survey,ross2011reduction} is an efficient way to transfer such knowledge to a robot, wherein the human expert teaches a robot by showing it instances of execution of the task of interest. There are numerous examples of fairly sophisticated behaviours being learnt by robots, such as for control and planning in high dimensional systems \citep{levine2016end,finn2017deep,gu2017deep}. A common theme across a majority of these works is that the target of transfer is the specific behaviour, so that the robot is taught to mimic a motion (with learning methods being used to generalise over, say, differing body configurations). We are also interested in being able to transfer other aspects of knowledge about the world which may allow the robot to infer deeper concepts. A necessary first step towards this form of teaching or transfer is to give the robotic agent the ability to learn models that represent structure in ways that are similar to corresponding human notions - so that it may be possible for the human expert and robot to leverage common grounding \cite{clark1985language}.  

From a different direction, there is recognition among roboticists that models and policies used by autonomous robotic agents should be explainable \cite{xai_baa, Wachtereaan6080}. The notion of what it means to interpret and explain is in itself a topic of active debate at the moment, but we believe it is safe to claim that grounded models used by robots should satisfy key desiderata. For instance,  the notion of similarity with which items are grouped together must align with corresponding human notions or that factors of variation implied by the model should be understandable to a human expert. These desiderata are also closely aligned with the emerging literature on learning disentangled representations \cite{higgins2016beta,chen2016infogan,chen2018isolating,burgess2018understanding}. 

In this paper, we focus on learning models of objects of the kind that robots might encounter in manipulation settings. In order to make the underlying concepts fully clear, we use simple objects that uncontroversially expose the structure of the domain. For instance, we show that after being exposed only to a few instances of a ball and a cube, the agent can figure out that other instances of cubes can be labelled as such (without this fact being explicitly stated by the expert). 

% In open environments, there are other forms of variability that frequently arise. The agent should be able to pick up variability both in the symbols used---what one calls sky blue, another might call turquoise---and in the groundings of those symbols---blue itself has multiple hues and shades. Often, in the same environmental set-up different tasks would require different discriminative abilities which are communicated through the labels the expert uses.

\begin{figure}
\centering
\includegraphics[scale=0.53]{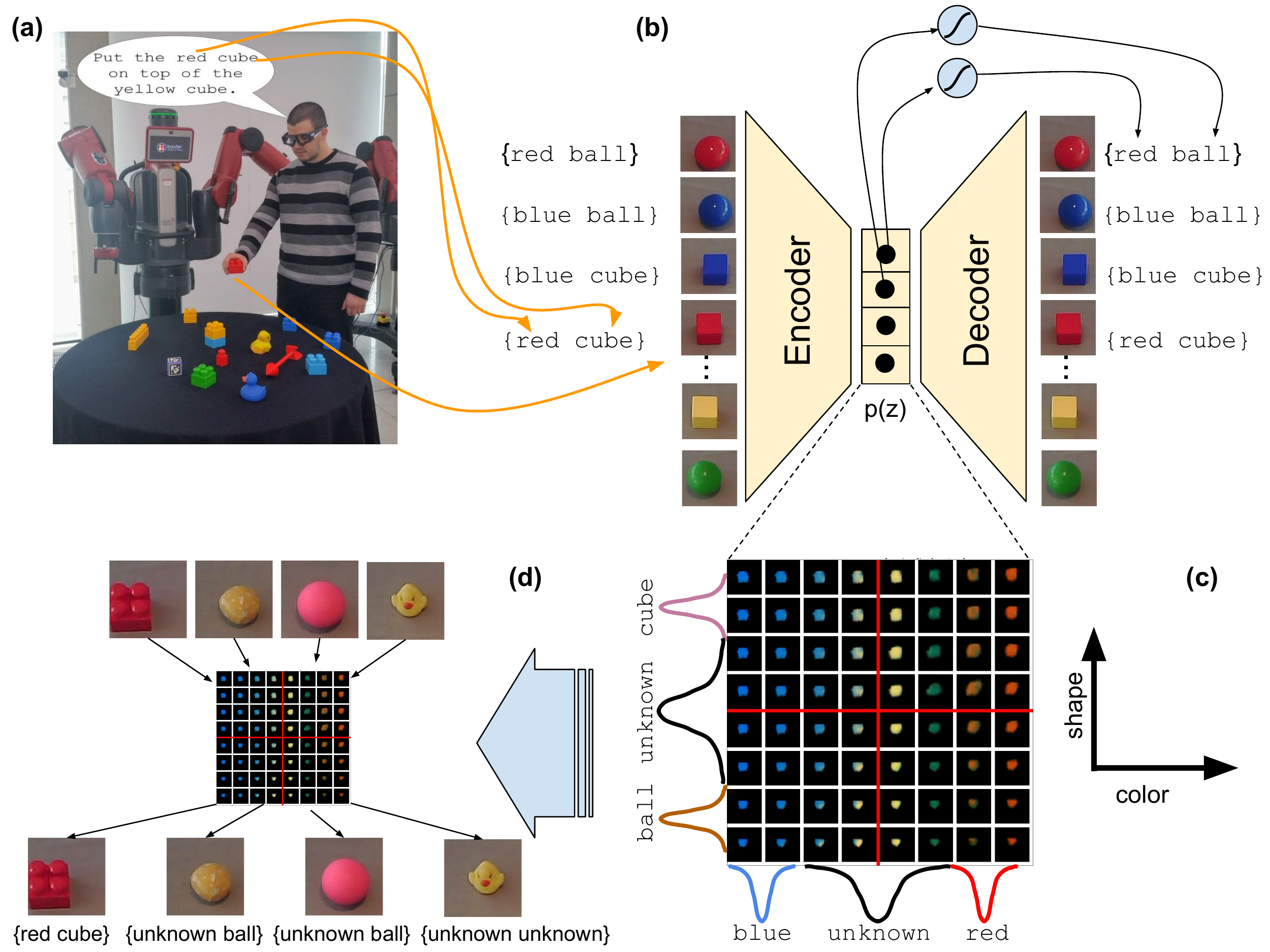}
\caption{Overview of the full framework. The expert is demonstrating the task at hand while uttering symbols which are grounded according to their attention in the environment. The conceptual grouping of the symbols is given. \textbf{(a)} The gathered dataset is used to train a variational autoencoder with a set of auxiliary classification losses - one per conceptual group. Each classifier takes information from a single latent dimension. \textbf{(b)} The training procedure guarantees that the semantic meaning of each group will be encoded in a separate latent dimension with linearly separable classes \textbf{(c)} Using the per-class estimated 1D Normal distributions, together with in-between distributions for \texttt{unknown} labels, we can perform 1-NN classification \textbf{(d)}}
\label{fig:pipeline}
\end{figure}

Our main contribution consists of a framework---see figure \ref{fig:pipeline}---which allows for independent user-defined factors of variation, manifested in a high-dimensional space, to be projected to a lower-dimensional latent space by preserving the factors' orthogonality. The latter is guaranteed by aligning each of the basis vectors that span the latent space with a single factor. Each factor is specified as a set of weak labels over the high-dimensional space. So, for the same data the framework can learn different representations for different sets of symbols. We test the framework both on a synthetic dataset (modified dSprites \citep{dsprites17} with added color), with controlled factors of variation and on a dataset of real-world objects captured from a set of human demonstrations for the task of sorting table-top objects according to the user's preference.

\vspace{-1em}

%===============================================================================

\section{Related Work}
% We present a method that combines grounded language acquisition \citep{kollar2013toward,matuszek2014learning,sinapov2011object,sinapov2013grounded,sinapov2014grounding} and physical symbol grounding \citep{vogt2002physical,harnad1990symbol}, motivated by the problem of robust human-robot interaction. 
An efficient, and unobtrusive, learning process must need minimal effort on the part of the human expert, requiring the agent to be able to generalise to unobserved scenarios. At the same time, the models and concepts used by the agent must be interpretable by the human, in order for the interaction to be efficient. $0$-shot, $1$-shot \citep{goodfellow2016deep} and meta-learning \citep{thrun2012learning} are all approaches which strive for fast learning and better generalisation from limited amounts of data. However, from the point of view of ensuring safe operation by a robot, it is also important to be able to quantify when an unfamiliar situation occurs and to seek guidance from an expert.

The ability to deal with examples that fall outside the domain of the training distribution is known as `open world recognition' \citep{bendale2015towards,lakkaraju2017identifying}. This is needed when an agent comes across \textit{unknown unknowns} - observations that have not been seen during training. In this setting, the incomplete nature of the agent's model causes it to make the wrong prediction. A complete model would cope with such cases by explicitly recognising that data point as `unknown'. It is desirable for the agent to be able to express its \textit{ignorance} to the expert in a way that is comprehensible to them. This motivates the need for interpretability - we learn a representation that enables communication for user feedback regarding partially labelled sensory observations. 

In the context of visual sensory inputs, `inverse graphics' is an approach to learning compressed and factorised representations of high-dimensional visual data - e.g. learn to invert a graphical renderer and infer (for a given image) the parameter vector that generated an input image. The fact that such vectors are low-dimensional and factorised is what makes them appealing from a human-interpretability point of view \citep{li2017deep,kulkarni2015deep}. 

In the representation learning literature, there has been work \citep{higgins2016beta,chen2016infogan,chen2018isolating,ainsworth2018interpretable,denton2017unsupervised} aimed at unsupervised learning of inverse-graphics-like representations of image datasets. This has been demonstrated with models like InfoGAN \cite{chen2016infogan} which does learn the desired factorised representations in an unsupervised fashion but also suffers from problems like unstable training and sensitivity to the chosen prior distribution over the latent codes. Moreover, there is no explicit inference module that would produce a latent code, given an observation, even though the recognition network Q can be used as such. An alternative in the literature is the $\beta$-VAE model \cite{higgins2016beta} which has fewer parameters to tune, has more stable training and reportedly learns similar representations. The factors of variation which are encoded in these representations are the ones that best explain the visual variations in the data \citep{burgess2018understanding} but might not necessarily be the factors of variation which are crucial to recognise and discriminate in the context of a particular task. Inferring which learnt factors best ground the user-defined symbols is usually a separate process wherein inference must be performed for each label-factor pair. Learning a representation suitable for symbol grounding \citep{vogt2002physical,harnad1990symbol} from raw pixels in an unsupervised manner, based on $\beta$-VAE, has turned out to be non-trivial \citep{higgins2017scan}. In this work we employ methods based on deep variational inference \citep{kingma2013auto} and perform the learning of a factorised vector space and the grounding of symbols in it simultaneously, in a weakly-supervised fashion. Thus we achieve better alignment between the user-defined semantic conceptual groups and the basis vectors of the vector space.

%===============================================================================

\section{Background and Problem Formulation}
\label{sec:formulation}

In this work, we assume that the human expert and robotic agent share the environment and that they can both observe and interact with objects in the surrounding world. The agent can extract images of distinct objects from the world and receive labels for them from the expert.

Data capture can be achieved with any technique that is able to align a sequential abstract plan for a task (parsed from a linguistic construct) with a temporal trace of expert movements and actions in the external environment - e.g. gesture recognition \citep{eldon2016interpreting}, movement tracking \citep{al2017natural}, eye-tracking \citep{penkovICRA17}. Thus, specific symbols from the plan can be related to regions of interest in the environment, where the object, encoding the label's meaning, resides.

Let $\mathcal{O} = \{\mathbf{x_1}, \ldots, \mathbf{x}_N\}, \mathbf{x}_i \in \mathbb{R}^D$ be a set of unlabelled observations, which have been passively gathered prior to the expert demonstration. An expert demonstration introduces a set of $L$ conceptual groups $\mathcal{G}$ = $\{g_{1}, \ldots ,g_{L} \}$ (e.g. color, shape, size, etc.), where each group is a set of mutually exclusive discrete labels: $g_{i} = \{y_{1}^{i}, ..., y_{n_i}^{i}\}, n_i = |g_i|$ (e.g. color can be red, blue, yellow, etc.). The demonstration also provides labels, one drawn from each concept group, to $M$ of the observations: $\mathcal{O} = \{(\mathbf{x}_1, \mathbf{y}_1), \ldots, (\mathbf{x}_M, \mathbf{y}_M), (\mathbf{x}_{M+1}, \emptyset) \ldots (\mathbf{x}_{N}, \emptyset)\}$, $\mathbf{y}_j = \{y^{p}\}$, $p \in \{1, \ldots, L\}$, $y^p \in g_p$ is a set of labels which have been attached to the image $\mathbf{x}$ - one per concept group.

The task is to project each $\mathbf{x} \in \mathbb{R}^D$ into $\mathbf{z} \in \mathbb{R}^C$ , $C \ll D$, where the space of $\mathbb{R}^C$ possesses the following properties:

\textbf{Axes Alignment} - guarantees a one-to-one mapping from the concept groups $\mathcal{G}$, to the orthogonal basis vectors which span $\mathbb{R}^C$. This would guarantee that independent concepts in image space are kept independent in $\mathbb{R}^C$---e.g. color does not depend on the size, orientation or shape of an object.

\textbf{Intra-group Linear Separability} - the latent clusters in $\mathbb{R}^C$, corresponding to the labels in each concept group $g_{i}$, are linearly separable across the basis vector which has been aligned with that concept group. 

These two properties allow for $\mathbb{R}^C$ to be used as a feature space for performing probabilistic symbol grounding with the ability to recognise \textit{unknown} objects. For that we use 1-NN classification with 1D normal distributions (one per label) for each concept group in $\mathcal{G}$---see Figure \ref{fig:pipeline} (c) and (d).

%===============================================================================

\section{Methodology}

We explore the effects of adding an auxiliary classification loss to a Variational Autoencoder \citep{kingma2013auto} as a base architecture, specifically the $\beta$-VAE \citep{higgins2016beta}. Through weak supervision, in the form of partially labelled data, the auxiliary loss influences the latent space of the model to exhibit properties which make it suitable for robust symbol classification. The model consists of a convolutional encoder network $q_\phi$, parametrised by $\boldsymbol{\phi}$, a deconvolutional decoder network $p_\theta$, parametrised by $\boldsymbol{\theta}$, and a set of linear classifiers parametrised by $\mathbf{w}_i \in \mathbb{R}^{|g_i|}$ for each group $g_i \in \mathcal{G}$. Additional parameters---$\alpha, \beta, \gamma$---are added on the three terms of the overall loss function---see (\ref{equation:loss})---in order to leverage their importance.
\begin{equation}
\min_{\theta, \phi, \mathbf{W}} \mathcal{L}(\mathbf{x}, \mathbf{y}, \theta,\phi) = \beta D_{KL} (q_{\phi}(\mathbf{z}|\mathbf{x})||p_{\theta}(\mathbf{z})) - \alpha \mathbb{E}_{q_{\phi} (\mathbf{z}|\mathbf{x})} (\log p_{\theta} (\mathbf{x}|\mathbf{z})) + \gamma \sum_{i}^{|\mathcal{G}|} H(z_{i}\mathbf{w}_i^T,\mathbf{y}_i)
\label{equation:loss}
\end{equation}
\textbf{Classification term} (weighted by $\boldsymbol{\gamma}$) - In order to force the learnt latent space to explain the variations in the data, we add a linear classifier for each concept group. Each classifier has to predict the set of labels for its assigned concept group using information only from a single latent dimension. No two classifiers have access to the same latent dimension. That forces each dimension to only explain labels from the particular concept group. A discrete cross-entropy term is used for the predictions of each classifier.

\textbf{Reconstruction term} (weighted by $\boldsymbol{\alpha}$) - The reconstruction loss is a standard Bernoulli Negative Log Likelihood, which is used to predict the pixel values across the three RGB channels. The motivation behind this term is that we do not assume all data points in our dataset are labelled. Thus the reconstruction loss would force data points which look similar in image space to be projected close to each other in the latent space.

\textbf{Kullback-Leibler divergence term} (weighted by $\boldsymbol{\beta}$) - The Kullback-Leibler divergence term ensures that the distribution of the latent projections of the data in $\mathbb{R}^C$ does not diverge from a prior isotropic normal distribution. A perfectly optimised KL term would result in all latent projections to be 0. This forces the encoder network $q_\phi$ to be more efficient when encoding the image observations so that their latent projection can be discriminated from each other across the basis vectors aligned with $\mathcal{G}$---the classification term---and the decoder network $p_\theta$ can efficiently reconstruct them---the reconstruction term.

The values for all three coefficients are chosen empirically such that the values for all the loss terms have similar magnitude and thus none of them overwhelms the gradient updates while training.

In order to account for data-generative factors of variation that might not be needed to encode the conceptual groups $\mathcal{G}$ but are still essential for good reconstruction and stable training, we allow $|\mathbb{R}^C| > |\mathcal{G}|$. For example, spatial and rotational factors of variation would not contribute to explaining the semantics of concepts like shape and size but should still be accounted for in order for the subset of basis vectors in $\mathbb{R}^C$ which are aligned with $\mathcal{G}$ to encode only information which explain $\mathcal{G}$. 

Algorithm \ref{alg:learning} describes the core functionality of the framework. Initially we have a set of partially labelled observations, a set of conceptual groups with their labels, and a single untrained linear classifier for each group. In the training process, each observed image $\mathbf{x}$ is passed through the network and its reconstruction $\hat{\mathbf{x}} \sim p_{\theta}(\mathbf{x}|\mathbf{z}), \mathbf{z} \sim q_{\phi}(\mathbf{x})$ is fed to the loss $\mathcal{L}$. If the data point is labelled, a label is predicted for each conceptual group from the respective classifier. After training, we estimate the parameters of a 1D normal distribution for each label across the dimension which was responsible for predicting it during learning.

\begin{algorithm}[h]
\caption{Model Learning with Weak Supervision}
\label{alg:learning}
	\KwIn{observations $\mathcal{O} = \{(\mathbf{x}_1, \mathbf{y}_1), \ldots, (\mathbf{x}_M, \mathbf{y}_M), (\mathbf{x}_{M+1}, \emptyset) \ldots (\mathbf{x}_{N}, \emptyset)\}$}
    \KwIn{conceptual groups $\mathcal{G}$ = $\{g_{1}, \ldots ,g_{L} \}$}
    \KwIn{linear classifiers $\mathcal{W} = \{\mathbf{w}_i\}, i \in \{1, \ldots, L\}$}
    \KwIn{Isotropic Normal Prior $p(z) = \mathcal{N}(0,I)$}
    \KwOut{set of per-label estimated 1D normal distribution for each label in each conceptual group: $K = \{\{\mathcal{N}(\mu_{q}^p,\sigma_{q}^p)\}, p \in \{1, \ldots, L\}, q \in \{1, \ldots, |g_p|\}$} \BlankLine
    \While{not converged}{
      \For{each $(\mathbf{x}, \mathbf{y})$ in $\mathcal{O}$}{
          $\boldsymbol{\hat{\mu}}, \boldsymbol{\hat{\sigma}} \leftarrow Encode(\mathbf{x})$\;
          $\mathbf{\hat{z}} \sim \mathcal{N}(\boldsymbol{\hat{\mu}}, \boldsymbol{\hat{\sigma}}I)$\;
          \If{$\mathbf{y} \neq \emptyset$}{
              \For{each $\mathbf{w}_i$ in $\mathcal{W}$}{
                $\hat{\mathbf{y}}^{i} \leftarrow z_{i} \mathbf{w}_{i}^{T}$\;
              }
          }
          $\mathbf{\hat{x}} \leftarrow Decode(\hat{\mathbf{z}})$\;
          Use $\hat{\mathbf{x}}$, $\hat{\mathbf{y}}$, $\boldsymbol{\hat{\mu}}$ and $\boldsymbol{\hat{\sigma}}$ to compute $\mathcal{L}$---see (\ref{equation:loss})\;
      }
    }
    \BlankLine
    \For{each $g_i$ in $\mathcal{G}$}{
    	\For{each $y_j^i$ in $g_i$, $j \in \{1,\ldots,|g_i|\}$}{
    		$\mathbf{f} \leftarrow \{(\mathbf{x}, \mathbf{y}) \in \mathcal{O} | y_j^i \in  \mathbf{y} \}$\;
            $\boldsymbol{\hat{\mu}}, \boldsymbol{\hat{\sigma}} \leftarrow Encode(\mathbf{f})$\;
          	$\mathbf{\hat{z}} \sim \mathcal{N}(\boldsymbol{\hat{\mu}}, \boldsymbol{\hat{\sigma}}I)$\;
            $\mathcal{N}(\mu_j^i,\sigma_j^i) \leftarrow fitNormal(\hat{\mathbf{z}}_i)$\;
            Add $\mathcal{N}(\mu_j^i,\sigma_j^i)$ to $K$\;
    	}		
    }
\end{algorithm}

When classification is performed at test time, the task is to predict a set of $L$ labels $\mathbf{y}$ for each image observation $\mathbf{x}$. Classification is performed by using the factored projection of each image observation in $\mathbb{R}^C$. The 1D coordinates along each basis vector $\mathbf{z}_i$ of $\mathbb{R}^C$ are used to predict a label for the correspondingly aligned concept group $g_i$. Each 1D value is normalised with respect to the normal distributions in $K$ corresponding to $g_i$ and the class associated with the closest one, along $\mathbf{z}_i$ is assigned to $\mathbf{x}$ for $g_i$. As a consequence of optimising the Kullback-Leibler divergence term, together with the reconstruction and classification losses, any data points that have not been labelled and do not resemble the labelled ones end up being projected closer towards the origin of the latent space $\mathbb{R}^C$, in between the clusters associated with the labelled data. Thus, in order to be able to account for such unknown objects, for every two neighbouring distributions $\mathcal{N}(\mu_{l}^i,\sigma_{l}^i)$ and $\mathcal{N}(\mu_{r}^i,\sigma_{r}^i)$, along the basis vector $\mathbf{z}_i$, aligned with $g_i$, we fit an average distribution $\mathcal{N}_{u}(\mu_{u}^i,\sigma_{u}^i)$ where $\mu_{u}^i = \frac{\mu_l^i + \mu_r^i}{2}$ and $\sigma_{u}^i = \frac{\sigma_l^i + \sigma_r^i}{2}$. Any data point that is closer to such an \textit{unknown} distribution than to a labelled one is considered \textit{unknown}.

%===============================================================================

\section{Experiments}

\subsection{Data}

\begin{figure}[h]
\centering
\includegraphics[scale=0.4]{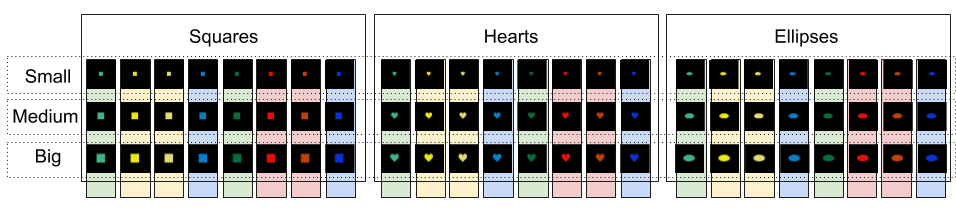}
\caption{A modified version of the dSprites \citep{dsprites17} dataset with added color. The RGB equivalents of 8 colors---two variations of red, green, blue and yellow---are added. Spatial factors of variation are also present in the images but not shown in the figure.}
\label{fig:data_sprites}
\end{figure}

The controlled data-generative factors of variation of the modified dSprites dataset---see figure \ref{fig:data_sprites}---make it suitable for exploring how the two baselines compare to the proposed model with respect to the defined manifold properties in section \ref{sec:formulation}. The resulting dataset is of size 3500 images---72 objects with spatial x/y variations in the image. We perform two experiments with the same underlying dataset but different sets of symbols in order to demonstrate how the user's preference is encoded in the latent space.

\begin{figure}
\centering
\includegraphics[scale=0.45]{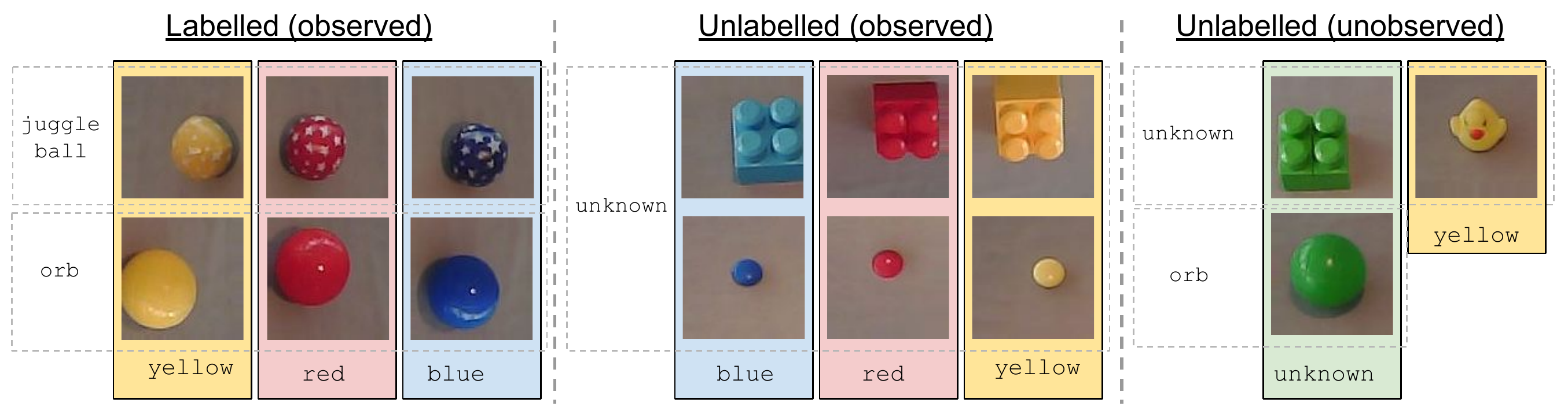}
\caption{Example images crops of all table-top objects. Spatial and pixel noise factors of variation were added to each crop through data augmentation techniques.}
\label{fig:data_real}
\end{figure}

In order to demonstrate the application of the framework to real-world human-robot interaction scenarios, a second dataset of objects on a table-top is gathered from a human demonstration - Figure \ref{fig:data_real}. The task the human performs is to separate the objects by their function - \textit{juggling balls} vs \textit{orbs}, and then by their color - \textit{red} vs \textit{yellow} vs \textit{blue}. Lego blocks and whiteboard pins are also present in the scene, but they are not manipulated and no label information is given about them from the expert. At test time the agent has to repeat the task, with new objects being present in the scene that were previously unobserved---\textit{green} objects and a \textit{yellow rubber duck}. Each object image is augmented, resulting in a dataset of size 7500 images of 15 objects with spatial variation. Both datasets are split into a training-testing sets with an 80-20 ratio. All results are reported on the test set. The setup for gathering and labelling per-object image crops in a single human demonstration through the use of eye-tracking is described in Appendix \ref{appendix:demo}.

\subsection{Evaluation}

Two baselines are used to benchmark against the proposed architecture - a Vanilla Beta-VAE ($\gamma=0$) and a Convolutional Classifier Network with Kullback-Leibler divergence and no reconstruction term ($\alpha=0$). More details on the network architecture and training setups are provided in Appendices \ref{appendix:architecture} and \ref{appendix:hyperparameters}. In order to evaluate the extent to which each baseline satisfies the properties we define in Section \ref{sec:formulation}, we use metrics which are inspired by the literature on learning disentangled representations \citep{eastwood2018framework}

\textbf{Axes Alignment} - In order to determine how well the concept groups in $\mathcal{G}$ are aligned with the basis vectors that span the latent space $\mathbb{R}^C$ we perform PCA on the latent projections of the data points for each label and examine the alignment of the resultant eigenvectors with the basis vectors. If a particular concept group is aligned with a single basis vector, then the eigenvector with the smallest eigenvalue should be parallel to that basis vector. Such alignment would mean that traversing a single concept group in image space corresponds to perturbing the values of a single basis vector. For example, if $\mathbf{z}_i$ encodes the concept of \textit{color}, then the latent distribution for all \texttt{blue} datapoints should have small variance along $\mathbf{z}_i$ and large variance across all other $\mathbf{z}_{j \neq i}$. To examine this, for each label in each concept group, we plot the cosine similarity diagrams depicting the cosine distance between each pair of basis and eigenvectors---see figure \ref{fig:axes_alignment}. White cells mark the cosine similarities between the shortest eigenvector with all basis vectors and between its closest parallel basis vector and all other principal components. The average entropy of the normalised white cosine values is reported for each combination of model-experiment to ease quantitative analysis. Low entropy corresponds to axis alignment, resulting in a single big white cell (for the smallest eigenvector and its most-parallel basis vector) and remaining small white cells along the row and column of the big one. For each concept group in each experiment we should observe the same white cell patterns if axis alignment is being achieved. For further information on this point, please see the supplementary materials \footnote{https://sites.google.com/view/interpretable-latent-spaces/axes-align-explain}.

\textbf{Intra-group Linear Separability} - In order to perform 1-NN classification at test time with unlabelled objects, using 1D Normal distributions across the concept groups in $\mathcal{G}$, the latent cluster in $\mathbb{R}^C$ for each label $y_j^i \in g_i$ has to be linearly separable from the latent clusters for all other labels from $g_i$. We report F1 scores for each class label, per concept group, including predictions for unlabelled observations which represent both \textit{known} and \textit{unknown} labels. 

\subsection{Results}
\label{sec:results}

\textbf{Experiment 1} - learn $\mathbf{z_0} \equiv $ color and $\mathbf{z_1} \equiv $ size. The user-uttered labels for color are {\textit{red, blue}} and for size are {\textit{small, big}}. The color labels are assigned to a single variation of the respective color. All images which can not be described by the given labels---\textit{yellow} and \textit{green} for color and \textit{medium} for size---are given an \textit{unknown} ground truth label. Total $|Z|$ = 4.

\textbf{Experiment 2} - learn $\mathbf{z_0} \equiv $ shape and $\mathbf{z_1} \equiv $ size. The user-uttered labels for shape are {\textit{square, heart}} and for size are {\textit{small, big}}. All images which can not be described by the user-uttered labels---\textit{ellipse} for shape and \textit{medium} for size---are given an \textit{unknown} ground truth label. Total $|Z|$ = 4.

\textbf{Experiment 3}  learn $\mathbf{z_0} \equiv $ color and $\mathbf{z_1} \equiv $ object type. The user-uttered labels for color are {\textit{red, blue}} and for object type are {\textit{juggle ball, orb}}. All images which can not be described by the user-uttered labels---\textit{lego bricks} and \textit{whiteboard pins, rubber ducks} or \textit{green objects}---are given an \textit{unknown} ground truth label. Total $|Z|$ = 4.

\begin{figure}[h]
\centering
\includegraphics[scale=0.55]{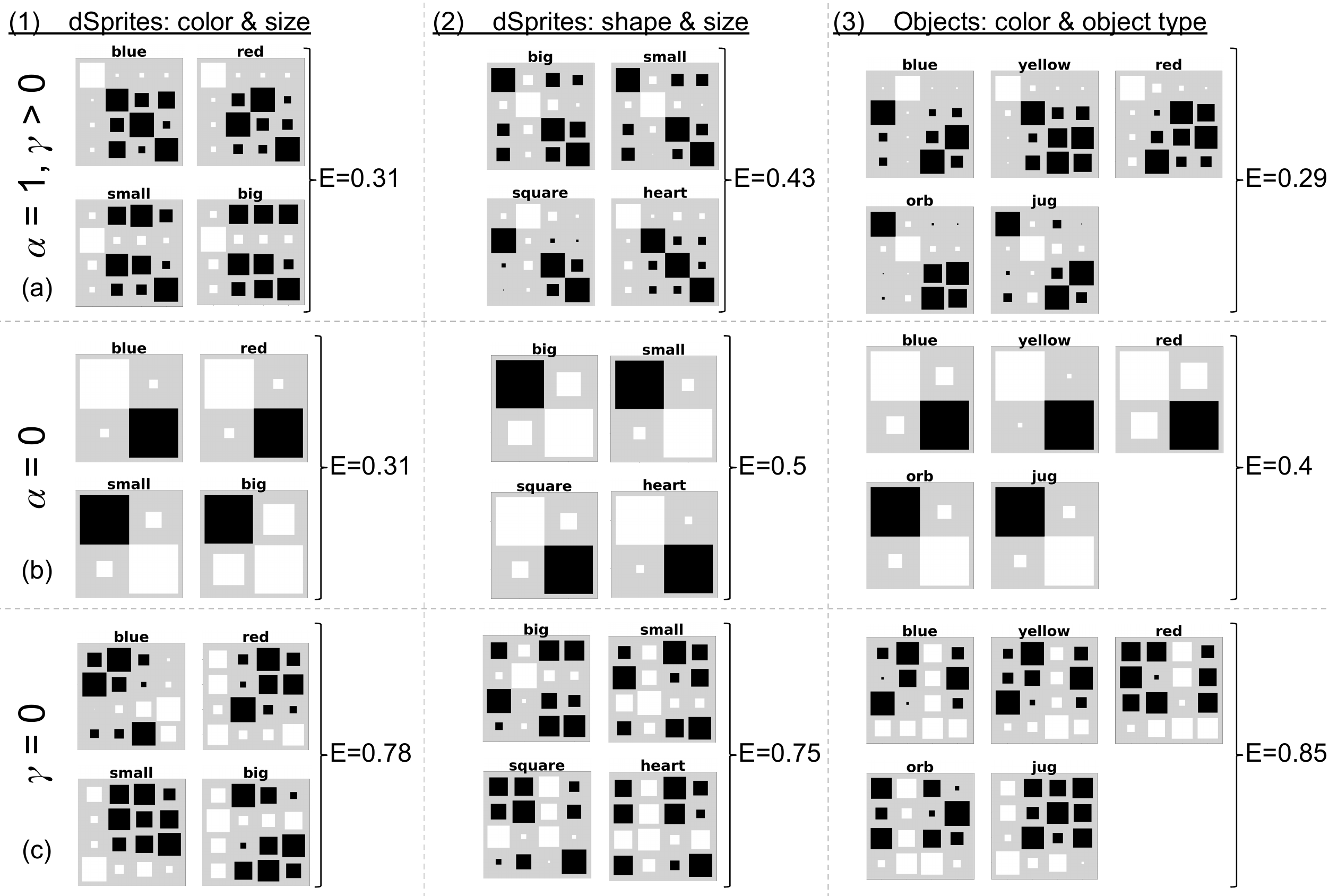}
\caption{Axes Alignment evaluation for the full model \textbf{(a)}, the classifier baseline \textbf{(b)} and the $\beta$-VAE \textbf{(c)} for the 3 experiments - \textbf{(1)}, \textbf{(2)} and \textbf{(3)} respectively. For each experiment and each model, the cosine similarity diagram for each label is shown. In each diagram the size of square (i,j) represents the cosine distance between eigenvector $\mathbf{c}_j$ and basis vector $\mathbf{z}_i$ in $\mathbb{R}^C$, after performing PCA on the latent values for the corresponding label. All white cells mark the cosine similarities between the smallest eigenvector with all basis vectors and between its most-parallel basis vector and all other principal components. $E$ denotes the average entropy estimates over normalised white cells values for a single model and a single experiment.}
\label{fig:axes_alignment}
\end{figure}

Figure \ref{fig:axes_alignment} presents the main findings from our experiments. Both the full model and the classifier network - figure \ref{fig:axes_alignment} (a,b) achieve good alignment between the concept groups in $\mathcal{G}$ and the basis vectors of $\mathbb{R}^C$. All labels from a particular concept group are consistently explained by the latent dimension which was used to predict them - e.g. color is explained by $\mathbf{z}_0$ and size is explained by $\mathbf{z}_1$ for experiment 1. For the classifier network - figure \ref{fig:axes_alignment} (b) we show only the first two latent dimensions which were used to perform the classification (hence containing relevant information). The $\beta$-VAE - figure \ref{fig:axes_alignment} (c) fails to satisfy the axis alignment requirement - labels from the same concept groups are best explained by different basis vectors---e.g. color in experiment 1. For this model we deterministically chose to use $z_i$ to predict labels for $g_i$ - Table \ref{table:f1}. This is of course owing to the fact that the $\beta$-VAE baseline is trained in a completely unsupervised fashion. We will report this to bring out the observation that non-axis-aligned representations render the 1-NN classification process to be non-trivial, as it is not at all clear which axes should be used to decide on the labels for each concept group. The full model and the classification network baseline perform comparably well with respect to classifying labelled data points. However, the classifier-only baseline is not as good at discriminating between known and unknown objects - \textit{unknown} columns in Tables \ref{tab:1}, \ref{tab:2}, \ref{tab:3}. We speculate that this is a result of the model having access only to labelled images during training, unlike the weakly trained full model. Thus, it is not able to pull together in the latent space both labelled and unlabelled visually similar data points in image space and push away visually different data points in image space. We provide more detailed analysis in the supplementary materials \footnote{https://sites.google.com/view/interpretable-latent-spaces/}.

\begin{table}[h]
    \begin{subtable}{1\linewidth}\centering
      {\begin{tabular}{ c|c c c||c c c}
      \hline
      Model & \textit{blue} & \textit{red} & \textit{unknown} & \textit{big} & \textit{small} & \textit{unknown} \\\hline
      $\alpha$ = 1, $\gamma \neq 0$ & \textbf{0.88} &\textbf{ 0.98} & \textbf{0.87} & 0.82 & \textbf{0.93} & \textbf{0.64}\\
      $\alpha$ = 0 & 0.76 & 0.78 & 0.29 & \textbf{0.84} & 0.87 & 0.52\\
      $\gamma$ = 0 & 0.05 & 0.37 & 0.23 & 0.45 & 0.34 & 0.26\\
      \end{tabular}}
      \caption{dSprites - $\mathbf{z_0} \equiv $ color and $\mathbf{z_1} \equiv $ size}\label{tab:1}
	\end{subtable}
	
   \begin{subtable}{1\linewidth}\centering
      {\begin{tabular}{ c|c c c||c c c}
        \hline
        Model & \textit{heart} & \textit{square} & \textit{unknown} & \textit{big} & \textit{small} & \textit{unknown} \\\hline
        $\alpha$ = 1, $\gamma \neq 0$ & 0.70 & 0.73 & 0.34 & \textbf{0.71} & \textbf{0.83} & \textbf{0.6}\\
        $\alpha$ = 0 & \textbf{0.97} & \textbf{0.8} & \textbf{0.67} & 0.53 & 0.79 & 0.5\\
        $\gamma$ = 0 & 0.2 & 0.33 & 0.32 & 0.4 & 0.38 & 0.33\\
    	\end{tabular}}
      \caption{dSprites - $\mathbf{z_0} \equiv $ shape and $\mathbf{z_1} \equiv $ size}\label{tab:2}
	\end{subtable}
   
   \begin{subtable}{1\linewidth}\centering
      {\begin{tabular}{ c|c c c c||c c c c}
      \hline
      Model & \textit{blue} & \textit{red} & \textit{yellow} & \textit{unknown} & \textit{juggle ball} & \textit{orb} & \textit{unknown} \\\hline
      $\alpha$ = 1, $\gamma \neq 0$  & \textbf{0.75} & \textbf{0.87} & 0.86 & \textbf{0.95} & \textbf{0.8} & 0.62 & \textbf{0.81}\\
      $\alpha$ = 0 & 0.7 & 0.31 & \textbf{0.89} & 0.1 & 0.64 & \textbf{0.64} & 0.41\\
      $\gamma$ = 0 & 0.2 & 0.25 & 0.25 & 0.18 & 0.1 & 0.3 & 0.26\\   
   	 \end{tabular}}
      \caption{Real objects - $\mathbf{z_0} \equiv $ color and $\mathbf{z_1} \equiv $ object type}\label{tab:3}
	\end{subtable}
    \caption{Evaluation of the discriminative abilities of each model for each experiment. F1 scores are reported for experiment 1 \textbf{(a)}, experiment 2 \textbf{(b)} and experiment 3 \textbf{(c)}.}
\label{table:f1}
\end{table}

\vspace{-1em}

%===============================================================================

\section{Conclusion}
We present a framework for physical symbol grounding where linguistically-defined semantic concepts from an expert, manifested in a high-dimensional image space, are mapped to a lower-dimensional learnt latent space. The resultant latent projections preserve any orthogonality between the user-defined concepts. In this sense, they are sufficient to perform robust---among other things, able to recognise {\textit{unknown}} unknowns---and sample-efficient symbol inference. We demonstrate this through experiments with images of computer-generated and real physical objects.

%===============================================================================
% The acknowledgments are automatically included only in the final version of the paper.
\acknowledgments{This work is partly supported by ERC Grant 269427 (STAC), a Xerox University Affairs Committee grant, and the CoGLE project under the DARPA XAI programme.}

%===============================================================================
\clearpage
\bibliography{bibliography}  % .bib

\clearpage

\begin{appendices}
\section{Demonstration and Object Crops Gathering}
\label{appendix:demo}

\begin{figure}[h]
\centering
\includegraphics[scale=0.55]{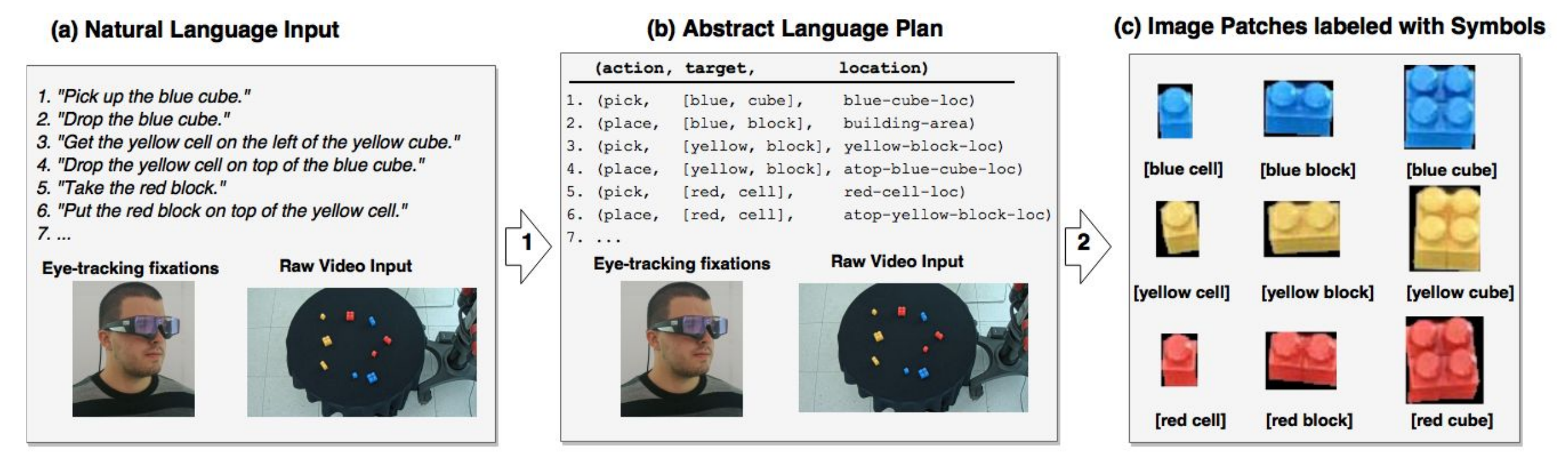}
\caption{Overview of the dataset capturing pipeline for an example task. Input to the system are natural language instructions, together with eye-tracking fixations and a camera view of the world from above \textbf{(a)} Natural language instructions are deterministically parsed to an abstract plan language \textbf{(b)} Using the abstract plan, a set of labelled image patches is produced from the eye-tracking and video data \textbf{(c)}}
\label{fig:etg}
\end{figure}

To capture the objects dataset with a single demonstration from a human we employ the pipeline depicted in Figure \ref{fig:etg} - the plan for the task is given in a natural language form which is deterministically parsed to an abstract sequential form of the type \verb!(action target location)!, where \verb!action! corresponds to an element from a predefined set $\mathcal{A}$, \verb!target! corresponds to a list of terms that describe an object in the world, and \verb!location! corresponds to a single symbol denoting a physical location in the environment. The narration of the plan execution by the human comprises of one sentence per abstract instruction. Therefore, given a plan description, our semantic parser finds a mapping from each sentence to a corresponding instruction, as defined by our abstract plan language. Elementary Dependency Structures (EDS) \cite{eds}, which are output by parsing the sentence with the wide-coverage English Resource Grammar \cite{erg}, are used as an intermediate step in this mapping procedure.

The demonstration is executed by the expert and their gaze fixations over the table surface are recorded for the duration of the demonstration at a rate of approximately 60Hz. The Pupil Labs eye tracking headset \cite{pupil} is used. Post-demonstration, the temporal alignment between the eye-tracking trace and the abstract sequential plan is performed using the GLIDE framework \citep{penkovICRA17}. For each fixation that is aligned to a particular instruction, a 100x100 pixel image, centered at the fixation, is cropped from the world-view camera of the headset. The \verb!target! labels from the instruction are attached to all corresponding crops. 

\section{Network Architecture}
\label{appendix:architecture}

The model architecture is implemented in the Chainer v4.3 framework\footnote{https://docs.chainer.org/en/stable/}. The network takes as input a set of RGB 100x100 pixel images and a set of labels and tries to reconstruct the same set of RGB 100x100 pixel images and predict the corresponding labels.

\begin{figure}[h]
\centering
\includegraphics[scale=0.30]{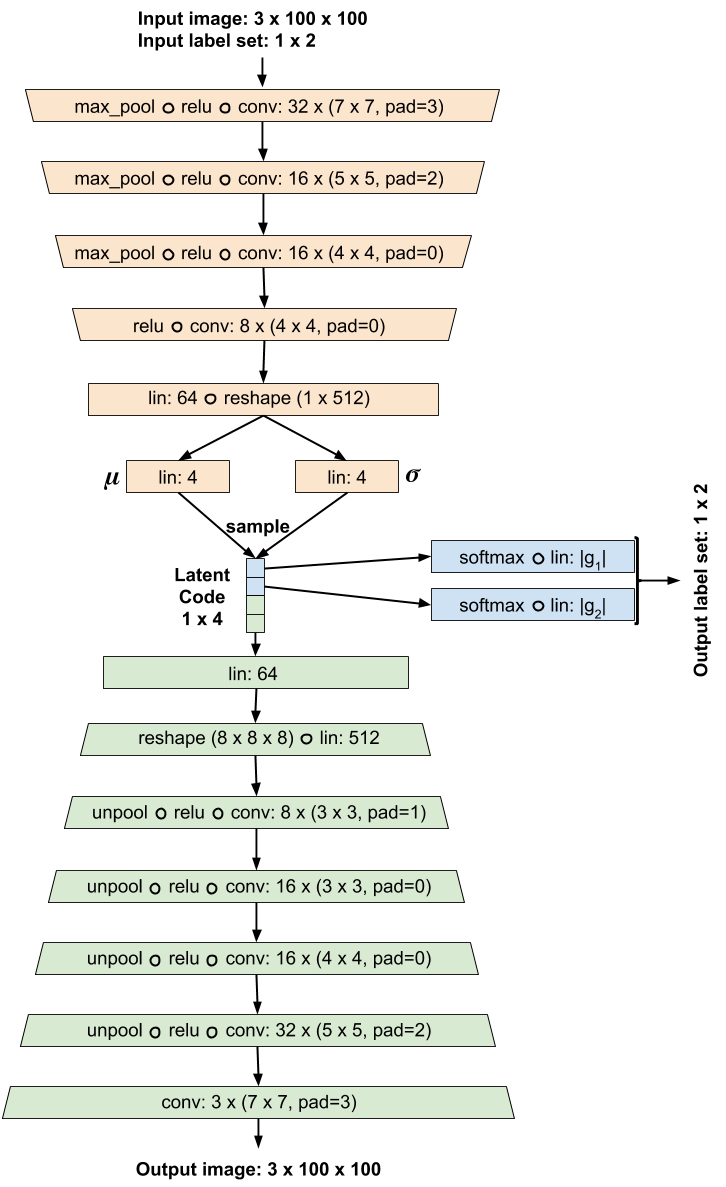}
\caption{Implementation details for the architecture of the trained models.}
\label{fig:model_architecture}
\end{figure}

For the encoding network---orange blocks in Figure \ref{fig:model_architecture}---a combination of standard convolutional and max-pooling layers is used. All max-pooling operations use a window size of 2x2. The output of each convolution is passed through a ReLU activation function. Before projecting into the latent space, all 8 8x8 pixel feature masks are unrolled and concatenated into a single 1D vector---1x512---which is then passed through two dense linear layers.

Once the distribution parameters---$\mu, \sigma$---for each data point are calculated, a single 1x4 latent code is sampled from each corresponding distribution.

For the decoding network---green in Figure \ref{fig:model_architecture}---the sampled latent code is passed through two dense linear layers and is then reshaped into 8 feature masks, 8x8 pixels each. Following that, a combination of standard convolutional layers and unpooling layers is used. All unpooling operations are performed with a window size of 2x2. The output of each convolution is passed through a ReLU activation function.

The outputs of the dense linear layers are not passed through any activation functions and all have bias terms. 

\section{Hyper parameters}
\label{appendix:hyperparameters}

Across all experiments, for all three models, training is performed for a fixed number of 200 epochs using a batch size of 32. The Adam optimizer \citep{kingma2014adam} is used through the learning process with the following values for its parameters---($learning rate=0.001, \beta1=0.9, \beta2=0.999, eps=1e-08, weight decay rate=0, amsgrad=False$)

For all experiments, the values (unless when set to 0) for the three coefficients from Equation \ref{equation:loss} are:
\begin{itemize}
\item Experiment 1 - $\alpha = 1, \beta = 30, \gamma = 10000$
\item Experiment 2 - $\alpha = 1, \beta = 30, \gamma = 10000$
\item Experiment 3 - $\alpha = 1, \beta = 10, \gamma = 10000$
\end{itemize}

The values are chosen empirically in a manner such that all the loss terms have similar magnitude and thus none of them overwhelms the gradient updates while training the full model.

\end{appendices}

\end{document}